\documentclass[conference]{IEEEtran}
\IEEEoverridecommandlockouts

\usepackage{cite}
\usepackage{amsmath,amssymb,amsfonts}
\usepackage{algorithmic}
\usepackage{graphicx}
\usepackage{textcomp}
\usepackage{xcolor}
\usepackage{array}

\def\BibTeX{{\rm B\kern-.05em{\sc i\kern-.025em b}\kern-.08em
    T\kern-.1667em\lower.7ex\hbox{E}\kern-.125emX}}
\begin{document}

\title{Event Classification of Accelerometer Data for Industrial Package Monitoring with Embedded Deep Learning}


\author{\IEEEauthorblockN{Manon Renault\IEEEauthorrefmark{1},
Hamoud Younes\IEEEauthorrefmark{2},
Hugo Tessier\IEEEauthorrefmark{3},
Ronan Le Roy\IEEEauthorrefmark{4},
Bastien Pasdeloup\IEEEauthorrefmark{1} and
Mathieu Léonardon\IEEEauthorrefmark{1}}
\IEEEauthorblockA{\IEEEauthorrefmark{1}
IMT Atlantique,
Brest, France 
}
\IEEEauthorblockA{\IEEEauthorrefmark{2}Imagination Technologies, United Kingdom}
\IEEEauthorblockA{\IEEEauthorrefmark{3} University of Toronto, Toronto, Canada}
\IEEEauthorblockA{\IEEEauthorrefmark{4}GoodFloow, Saint-Nazaire, France}
}

\maketitle

\begin{abstract}
Package monitoring is an important topic in industrial applications, with significant implications for operational efficiency and ecological sustainability. 
In this study, we propose an approach that employs an embedded system, placed on reusable packages, to detect their state (on a Forklift, in a Truck, or in an undetermined location).  
We aim to design a system with a lifespan of several years, corresponding to the lifespan of reusable packages.
Our analysis demonstrates that maximizing device lifespan requires minimizing wake time. 
We propose a pipeline that includes data processing, training, and evaluation of the deep learning model designed for imbalanced, multi-class time series data collected from an embedded sensor.
The method uses a one-dimensional Convolutional Neural Network architecture to classify accelerometer data from the IoT device. Before training, two data augmentation techniques are tested to solve the imbalance problem of the dataset: the Synthetic Minority Oversampling TEchnique and the ADAptive SYNthetic sampling approach. 
After training, compression techniques are implemented to have a small model size.
On the considered two-class problem, the methodology yields a precision of 94.54\% for the first class and 95.83\% for the second class, while compression techniques reduce the model size by a factor of four.
The trained model is deployed on the IoT device, where it operates with a power consumption of 316~mW during inference.

\end{abstract}


\section{Introduction}

In the current era of Industry 4.0, managing packages is a topic of interest in many industrial companies. Highlighted in~\cite{smart_packaging}, the Industrial Internet of Things (IIoT) is one of the challenges of smart packaging. It allows a transition from disposable to sustainable packaging, reducing the environmental impact of disposable cardboard.
As packaging is one of the biggest contributors to plastic waste and greenhouse gas emissions, reusable packaging has been an emerging, environment-friendly solution that aims at replacing cardboard boxes with Reusable Plastic Crates.
Monitoring package locations is a straightforward way to prevent loss, damage, or theft.

The location of packages could be determined using methods that use a Global Positioning System (GPS, as in~\cite{GPS}), Time Difference of Arrival (TDoA, as in~\cite{TDOA}), or Bluetooth Low Energy (BLE, as in~\cite{BLE}). These methods have a set of shortcomings that form a challenge when adopted for industrial package monitoring.
Using a GPS could be considered to monitor and track a package outside, but the tracking accuracy deteriorates once indoors or in harsh environments.
To apply TDoA, a dense network of IoT devices and antennas is required, with a high chance of interference in areas around city centers. As for BLE-based methods, they are useful for short-range tracking with multiple BLEs deployed in the desired area. Due to the lack of reliability (\emph{e.g.}, using GPS indoors), high operating cost (\emph{e.g.}, multiple antennas), and feasibility (\emph{e.g.}, placing BLEs in multiple factories that don't belong to the same company), real-time monitoring of packages is not possible. Thus, if a package is lost or damaged, it would not be feasible to determine the responsible party (factory or distributor). An alternative would be to track the package with human intervention, which is both time-consuming and costly. 

In summary, each of these solutions is relevant in different scenarios. This highlights the central premise of our proposal: the critical capability required in such IoT systems is the accurate identification of key events throughout the package's lifecycle. This study demonstrates that key events can be detected using accelerometer data and deep learning techniques.
The identified state changes can subsequently trigger additional processes, such as network communication, based on the package's status.
To the best of our knowledge, while some work on classification using accelerometer data is available \cite{sport_activity_classification} \cite{machinery_value_estimation}, no prior work on tracking using accelerometer signals has been addressed in the literature.

However, IoTs have hardware limitations: they have limited memory and limited computing capabilities. For instance, the study \cite{compsac_software_architecture} highlights the energy cost associated with deploying a neural network model directly on an IoT device. This is where Tiny Machine Learning (TinyML) plays a crucial role, bridging the gap between machine learning algorithms and hardware with limited computational capacity. It offers several advantages, including low energy consumption, low latency (inference is performed locally without the need to transfer data to a server), and enhanced security (data is stored locally on the device).
TinyML has already been adopted in several applications, such as agriculture~\cite{agriculture_tinyml}, healthcare~\cite{healthcare_tinyml}, or industrial automation~\cite{tinyml_survey}.

In this work, we show that embedded deep learning can solve the challenging task of package monitoring in the industry using data from an accelerometer.
To ensure a long lifespan, IoT devices must remain in a low-power sleep mode and wake only when necessary. This is achieved by waking the IoT device only when an event occurs and classifying it quickly, with low energy consumption.
This article presents an experiment focused on detecting two events: when the package is handled by a Forklift and when it is carried by a Truck.

The deep neural network deployed on the IoT device must maintain high precision to minimize classification errors. Such errors would require additional operations by the IoT device, thereby extending its wake time. This would directly impact the battery lifespan by reducing it. It must also have a high recall, as a classification failure requires a retry, which consumes additional energy.
The contribution of this paper is a methodology that suits our case, but can also be applied more broadly to neural networks dealing with imbalanced time series data and implemented on embedded devices that have a limited energy capacity.

The main contributions of this study are as follows:
\begin{itemize}
    \item We implement data augmentation on an imbalanced time series dataset. We test the Synthetic Minority Over-sampling TEchnique (SMOTE)~\cite{chawla_smote_2002} and the ADAptive SYNthetic Sampling (ADASYN)~\cite{adasyn} on the dataset and adapt them to our data for time coherence.
    \item We propose a Convolutional Neural Network (CNN) architecture and apply a threshold to the model's predictions to reduce energy consumption. If a prediction does not exceed this threshold, the data is automatically assigned to a special third class, indicating misclassification. This approach enhances prediction reliability and robustness.
    \item  Efficient compression techniques are applied to reduce the memory required for deploying the CNN on the ESP32 microcontroller, which typically supports the software stack of connected sensors in IoT devices. Our system, validated on the collected dataset, demonstrates that the embedded CNN can accurately detect and differentiate package location events (such as on a Forklift inside a factory or in a Truck during distribution) while maintaining low energy consumption.
\end{itemize}

\section{Methodology}


\subsection{Dataset and Misclassification Class}
\label{sec:dataset}
This methodology is applied to multi-class time series datasets.
A special class, referred to as "Dummy," is introduced to represent both irrelevant data for the IoT and data that are incorrectly classified.
This class represents scenarios such as a package falling to the ground, or a human manually moving it.
By categorizing such events as "Dummy", the IoT system can avoid unnecessary processing or misinterpretation, allowing it to remain energy-efficient and focused on meaningful events. This approach also reduces the risk of cascading errors, contributing to the robustness of the system in real-world applications.

\subsection{CNN architecture}
\begin{figure*}[btp]
\centering
    \includegraphics[width=0.7\linewidth]{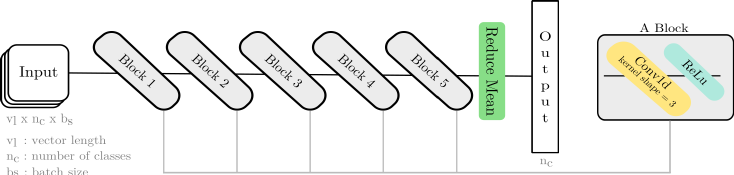}
    \caption{Detailed model architecture.}
    \label{fig:model_architecture}
\end{figure*}

The proposed deep learning model, described in Figure~\ref{fig:model_architecture}, is composed of one-dimensional convolutional layers followed by batch normalization. Normalization is implemented on activations of each layer across a mini-batch. It normalizes the activation values by computing the standard deviation of the mini-batch.
Rectified Linear Units (ReLU) activation layers are then applied.
These layers replace all negative input values with zeros.

The model expects an input of dimension ($V_{l}$, $l_{axes}$): $V_{l}$ corresponds to the time window size in samples, and $l_{axes}$ the number of axes of the accelerometer.
The vector length $V_{l}$ was chosen based on the value that yielded the highest precision (see Section~\ref{sec:experiments}).
Thus, the model's input is set following the vector length, giving an input of size 
$V_{l}$ x $l_{axes}$ and predicts outputs corresponding to the target classes, along with an additional class designated for misclassified data (see section~\ref{sec:dataset}).

Once the model has computed the classification, we pass the output data into a softmax function.
This function exponentiates each input vector component and then normalizes it by dividing by the sum of all the exponentiated components. This results in a valid probability distribution vector.

\subsection{Metrics} \label{metrics}

A threshold is applied to the maximum value of the model's predictions (see Section~\ref{threshold}). If the maximum value is greater than the threshold, the final classification is the value predicted by the model. Otherwise, the output is classified as "Dummy". Thus, the precision of the predicted value can be chosen.

Due to the dataset's significant class imbalance, accuracy is not a suitable performance metric. The imbalance can make the model disproportionately learn more from the majority class, thus skewing the accuracy metric to primarily reflect the proportion of the majority class, instead of the model's overall effectiveness across all classes.

Precision and recall metrics are more adapted to find the optimal threshold.
From these metrics, the optimal threshold can be found by plotting the precision-recall curve, as presented in Figure~\ref{fig:data_augmentation}.


\subsection{Thresholding soft decisions} \label{threshold}

The lifespan is a crucial topic in embedded devices, which have limited battery longevity.
Defining a threshold can generally benefit embedded machine learning, as the precision of the prediction can be chosen. In addition, it is particularly useful when a dataset is imbalanced. 

According to performance metrics, we can determine the best threshold value by finding a trade-off between precision and recall (see Section~\ref{metrics}). For the context considered, it is important for the prediction to be correct even if it necessitates allowing a few instances to be misclassified and get into the "Dummy" class.
In case of a bad prediction, some algorithms can be implemented to perform corrections. However, this produces a relatively important number of operations, hence high energy consumption.
To optimize classification performance, we select a threshold that maximizes precision. 

Figure~\ref{fig:data_augmentation} illustrates the different thresholds that can be selected and their impact on the model's performance.


\subsection{Data augmentation}
Imbalanced datasets are common in time series as collecting data often has constraints.
These imbalances can lead to misclassifications as the model has less minority data to learn with.
A common technique to address this issue is dataset re-sampling, which increases the representation of minority classes.
We test two of such data augmentation techniques, both based on interpolation: Synthetic Minority Oversampling TEchnique (SMOTE)~\cite{chawla_smote_2002}, and ADAptive SYNthetic sampling (ADASYN)~\cite{adasyn}.\\

\subsubsection{SMOTE algorithm}

Because our data are time series, we have to keep a continuity between each sample created, so we adapt the SMOTE algorithm to suit accelerometer data.
We generate a new vector \(x_{new}\), with the help of interpolation between two samples. To achieve this, we randomly choose a vector from the minority class, \(x_{selected}\), compute the Euclidean distance between other samples and choose the nearest neighbors.
One of the nearest neighbors is randomly selected, \(x_{neighbor}\), and linear interpolation is performed using a random coefficient $\alpha \in [0,1]$. This coefficient is multiplied by the difference between the vector we selected and the vector to interpolate with:
\[x_{new} = x_{selected} + \alpha * (x_{neighbor} - x_{selected})\]
This process is repeated based on the desired number of synthetic samples.\\

\subsubsection{ADASYN algorithm}

The main difference with SMOTE technique is that instead of applying a fixed coefficient to oversample uniformly minority data, it adapts the number of data to be resampled based on the degree of data imbalance. It focuses on minority data that are difficult to learn, by increasing minority data samples that are close to majority ones. This method helps the model to distinguish samples that are close to the decision boundary.

The algorithm is also adaptive, so we don't have to choose the number of synthetic samples we need to create.

The number of synthetic samples is determined by a ratio $r_i$, defined as $r_i = \frac{\Delta}{K}$, where $\Delta$ represents the number of samples within the $K$ nearest neighbors belonging to the majority class. This ratio $r_i$ is subsequently normalized to form a density distribution, $\hat{r_i}$. Each synthetic sample $g_i$ is then generated as:
$$
g_i = \hat{r_i} \times G \;,
$$
where $G$ is calculated as:
$$\label{eq:1}
G = (m_l - m_s) \times \beta \;,
$$
With $m_l$ the length of the majority class, $m_s$ the length of the minority class, and $\beta \in [0,1]$, a coefficient determining the balance level between the two classes.

\subsection{Compression techniques}

\textit{Deep compression}~\cite{deep_compression} is an efficient technique to reduce the size of the model while keeping its performance and improving computation efficiency. It consists of applying pruning, quantization, and Huffman coding on the model.
Once the model is trained, we convert it into an Open Neural Network Exchange format (ONNX)~\cite{onnx}, an open-source ecosystem. We use it as a tool to visualize the model architecture and to convert the PyTorch model to a TensorFlow Lite format~\cite{tflite}.
TFLite is a framework proposed by TensorFlow and optimized for machine learning on embedded systems.
Metrics scores were computed on the TFLite model. \\

\subsubsection{Pruning} 

As shown in~\cite{pruning}, pruning is a method useful for better generalization and better time inference. 
In~\cite{cnn_compression_microcontrollers}, it is shown that model size decreases as pruning levels increase for both floating-point and 8-bit quantized models.
In our work, we applied a weight rewinding technique from \cite{pruning_rewind}. This method consists of training the model with all its weights before pruning is applied. Then, the model is trained a second time on the remaining unpruned weights, reinforcing learning on these important weights.
During this retraining stage, the learning rate from the final epoch of the initial training is retained. We applied L1-norm unstructured pruning to the model weights. This approach eliminates weights with the lowest L1-norms by setting them to zero. By zeroing out these low-magnitude weights, we ensure that the pruning process minimally impacts the network's performance while reducing the computational burden. \\

\subsubsection{Quantization} 

As shown in~\cite{wu2016quantizedconvolutionalneuralnetworks}, it is possible to quantize convolutional neural networks to obtain a faster computation rate and to reduce the model size, without a significant loss in precision and recall of the predictions. 

Quantization is performed when converting the ONNX~\cite{onnx} format to TFLite~\cite{tflite}.
We explore one type of quantization adapted to the ESP32 microcontroller: \textit{full integer quantization}.
To perform this quantization, we initialize a converter function, referred to as the TFLite converter, that utilizes the INT8 quantization scheme. The converter requires a representative dataset, which is a small dataset used to calibrate the range (\emph{i.e.}, minimum and maximum) of all the floating point values in the baseline model. The representative dataset is composed of 100 samples chosen randomly from the training or validation dataset. Such a method ensures that both weights and activations are quantized for improved latency, processing, and power usage on integer-only hardware accelerators.
All weights, biases, and activation layers are converted from 32-byte floating-point to  8-byte integer format.
The input data is quantized at the beginning of the model.
Once the model has performed its implementations, the output data is converted back to 32-byte floating-point. \\

\subsubsection{Entropy encoding} 

Encoding is the process of converting data into a specialized format for efficient storage. One of the commonly used encoding categories is "lossless compression"~\cite{huffman}. Encoding methods in this category allow the recovery of the original data from the encoded data without any loss. Its basic idea is to encode data based on its frequency of appearance in a stream. Thus, it assigns lower and higher bits for the most and least frequent data, respectively, resulting in a high compression rate.  The encoding comes at the cost of negligible memory overhead for the "encoding dictionary" and increased latency for the decoding process.

Unstructured pruning is particularly effective when combined with entropy encoding, as it sets a large proportion of weights to zero, which can then be encoded using very few bits, significantly reducing the storage requirements. As a result, the entire neural network has a much smaller footprint, making it highly efficient in terms of memory usage. This synergy is particularly advantageous for deploying models on resource-constrained devices.

\section{Experiment} \label{sec:experiments}

\subsection{Dataset collection}

The dataset is composed of events recorded with an ultra-low-power module, consisting of a Micro-Electro Mechanical System (MEMS) accelerometer.
To collect data, four IoT devices were placed in a reusable package at different locations. Data were collected within a training facility for logistics technicians, where participants are instructed in operating trucks and forklifts. 
During the package lifecycle, events of the package are classified into three categories:

\begin{itemize}
    \item \textit{Forklift:} the package is transported and lifted via a Forklift from one place to another.
    \item \textit{Truck:} the package is placed at the back of a Truck. The latter could be moving or idle while the data was collected.
    \item \textit{Dummy:} this event describes all other movements that the package encountered (being moved by a person, opened, etc.), data that did not meet the threshold criteria (see Section~\ref{threshold}), and any misclassified data.
\end{itemize}
In total, the dataset is composed of 382,000~samples for Forklift, 2,520,000~samples for Truck, and 9,912~samples for Dummy. Samples were collected with the tri-axial accelerometer at a sampling frequency of 20~Hz.
Because the "Dummy" class represents misclassified samples and insignificant events, it was omitted from the performance results.
The dataset used in this study is not publicly available due to proprietary restrictions.
\subsection{Finding the optimal threshold}

The threshold is based on the performance of the model. As said in~\ref{threshold}, the highest the performance is, the lowest the energy the device will consume.
According to Figure~\ref{fig:data_augmentation}, we can find the best trade-off between precision and recall for each class.
When the model predicts a class with a confidence lower than the selected threshold, the data is directly classified into "Dummy" class, triggering verification methods that are energy efficient.
Thresholds were adapted to fit the optimal trade-off between precision and recall depending on the specific configurations being tested. Additionally, the threshold values were adjusted for each class based on the model's performance.

\subsection{Data augmentation}

For SMOTE and ADASYN algorithms, at every synthetic sample to generate we chose  5 nearest neighbors.
For ADASYN we chose $\beta$ equals to 0.4 in~\ref{eq:1}. For the SMOTE algorithm, we adapted the number of generated samples to be the same as the ADASYN one.

\begin{figure}[t]
    \includegraphics[width=1\linewidth]{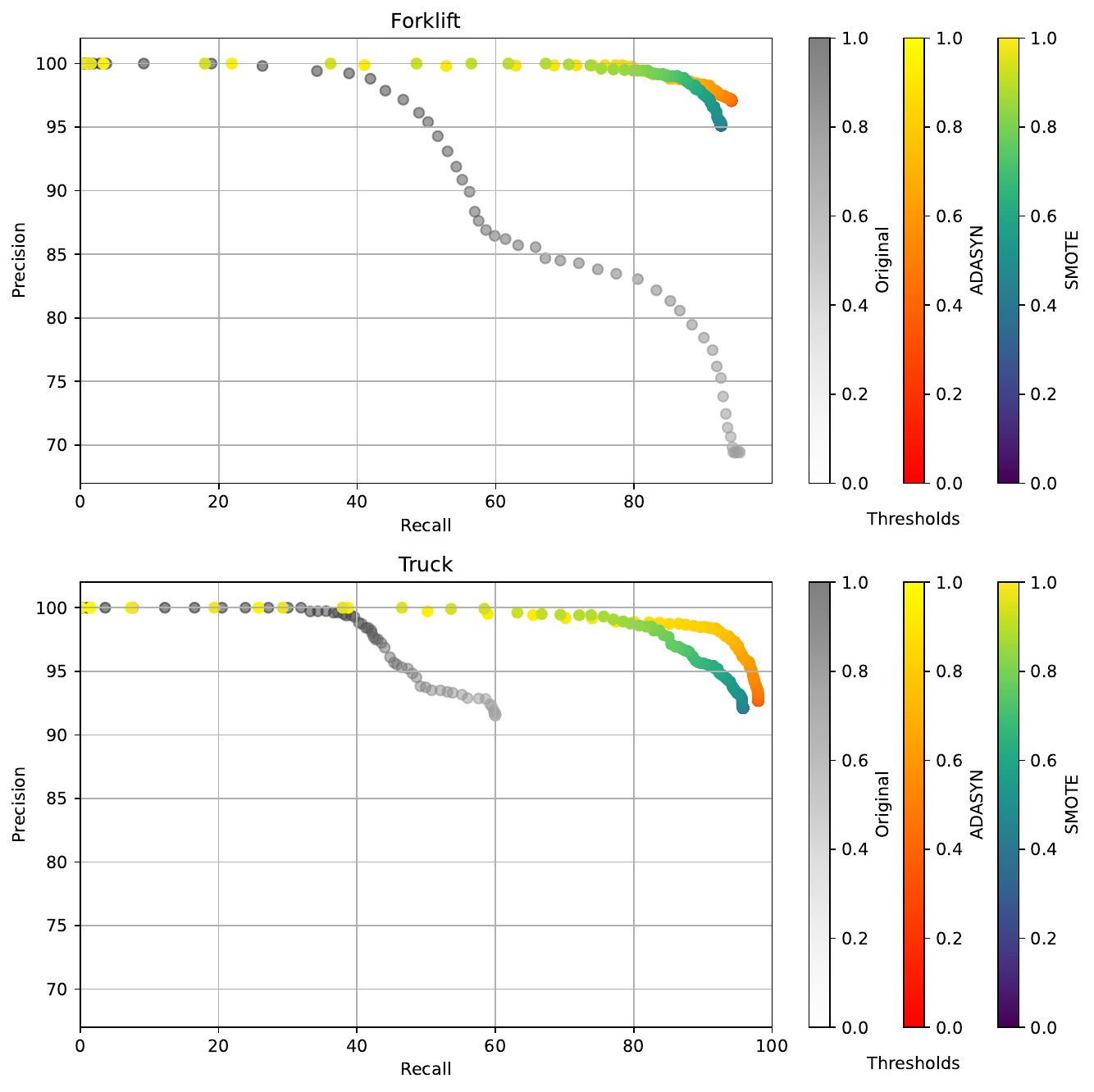}
    \centering
    \caption{Precision-recall curves with data augmentation techniques. Different curves for each technique are generated by varying the classification thresholds.
}
    \label{fig:data_augmentation}
\end{figure}

To evaluate the effectiveness of data augmentation, we compare the performance of SMOTE and ADASYN. As demonstrated with Table~\ref{tab:model_performance_threshold} and Figure~\ref{fig:data_augmentation}, we obtain a better performance with ADASYN technique.
This variation, as previously discussed, results from the sampling strategy employed by ADASYN, which prioritizes generating synthetic samples in areas close to decision boundaries between classes. By concentrating on these regions, ADASYN is particularly effective at enhancing the distinction of challenging and ambiguous cases. In contrast, SMOTE adopts a more uniform sampling approach across the dataset, which may fall short in addressing the class imbalances that occur specifically near decision boundaries, where the differentiation between classes is most critical.
With ADASYN algorithm, for predictions with a confidence score greater than the selected threshold, the model demonstrates a precision of 98.05\% for Forklift and 95.75\% for Truck.
Consequently, the ADASYN algorithm was selected for the next implementations.

\begin{table}[h]
    \centering
    \begin{tabular}{c|cc|cc}
         \textbf{Dataset configuration} & \multicolumn{2}{c}{\textbf{Forklift}} & \multicolumn{2}{c}{\textbf{Truck}} \\
          & Precision & Recall & Precision & Recall \\
         \hline 
         Original dataset & 82.17 & 83.21 & 91.56 & 59.98 \\
         \hline
         Dataset with SMOTE & 97.95 & 89.11 & 94.86 & 92.38 \\
         \hline 
         Dataset with ADASYN & 98.05 & 91.17 & 95.75 & 96.62\\
    \end{tabular}
    \vspace{3mm}
    \caption{Model performances across different data augmentation techniques.}
    \label{tab:model_performance_threshold}
\end{table}

\subsection{Training routine}

The CNN, referred to as the baseline model, has been designed using PyTorch and trained on an NVIDIA RTX3080 GPU. The event detection dataset has been split into 70\% training and 30\% for validation and testing. The training process has been performed for 20 epochs with a batch size of 200. 
We used Stochastic Gradient Descent as an optimizer. In order to make the learning rate more adaptive to the gradient descent optimization, we added a learning rate scheduler.
To reduce the noise of the labels, we added label smoothing to the Cross Entropy loss function and set it to 0.1. We also take into account the weights of the classes in the loss function.
For the optimizer, the learning rate is set to 0.5, the momentum to 0.9 and there is a weight decay set to 2e-5.


\subsection{Compression techniques}
\subsubsection{Performance}
Using the weight rewinding technique, the model undergoes two training phases: first, it is trained with all its weights; then, after pruning, the unpruned weights are fine-tuned to recover performance. Based on the study presented in~\cite{cnn_compression_microcontrollers}, a pruning ratio of 50\% was selected as an optimal trade-off between model compression and accuracy.

Pruning was applied to the weights of the convolutional and linear layers to reduce the memory footprint while maintaining acceptable model performance. The compression gain discussed in this work refers specifically to the model's footprint in non-volatile memory, which is limited due to hardware constraints and other software requirements, while dynamic memory is less constrained.
To further reduce memory footprint, Huffman coding is applied to the TFLite model after pruning. The impact of pruning and quantization on model size and performance is summarized in Table~\ref{tab:deep_compression_results}. Pruning reduces the model size by a factor of 1.5 while preserving high precision and recall values. This reduction is lower than a factor of 2, which would be anticipated with a pruning ratio of 50\%, because we compare the compressed versions of both the original and pruned models, rather than their raw sizes.

After quantization, the model size is reduced to 14~kB, achieving a compression factor of more than four compared to the original model. The combination of pruning, dynamic quantization, and Huffman coding effectively minimizes the model’s memory footprint while maintaining high precision, ensuring its feasibility within the stringent hardware constraints of the target deployment environment.\\

\begin{table}[h]
\centering
    \resizebox{\columnwidth}{!}{
        \begin{tabular}{>{\centering\arraybackslash}p{0.35\linewidth} >{\centering\arraybackslash}p{0.15\linewidth} p{0.1\linewidth} p{0.1\linewidth} p{0.1\linewidth} p{0.1\linewidth}}

            \textbf{Configuration} & \textbf{Compressed model size}  & 
            \multicolumn{2}{c}{\textbf{Forklift}} & \multicolumn{2}{c}{\textbf{Truck}}  \\
             &  &  Precision & Recall & Precision & Recall  \\
            \hline
            Original &  51~kB & 98.05 & 91.17 & 95.75 & 96.62 \\
            \hline
            Pruning & 33~kB   & 93.61 & 96.12 & 95.72 & 93.85 \\
            \hline
            Pruning + Quantization & 14~kB  & 94.54 & 92.59 & 95.83 & 88.66\\
            \end{tabular}
    }
    \vspace{1.5mm}
    \caption{Deep compression effect on precision, recall, and model size.}
\label{tab:deep_compression_results}
\end{table}

\subsubsection{Deployment on the microcontroller} 

From the TFLite model format, we convert the model to a CC code file (C++ Source Code File) before flashing it on the IoT device, as the microcontroller does not natively support the TFLite format.
The software Framework used on the ESP32 is the Epressif IoT Development Framework (ESP-IDF) (\cite{esp32}).

Although the ESP-32 module has 16 MB of flash memory, that is enough to fit any of the baseline models. Usually, this type of memory is shared with other modules in an IoT (such as WiFi, Bluetooth, etc.). Thus, the proposed compression scheme facilitates the efficient deployment of even deeper and more complex models. Moreover, as the quantized and pruned model size is relatively small, the model can be fit completely on the SRAM (Static Random Access Memory), which offers faster access compared to flash memory, thereby minimizing memory access overhead. \\

\subsubsection{Energy consumption} 

Reducing energy consumption extends the IoT device’s operational lifespan. Therefore, minimizing energy consumption is an important objective of the project.



In typical daily usage, each time the IoT device detects motion, it wakes up and performs an inference to determine if a state change has occurred. 
The energy consumed by each inference can be assessed using a typical power consumption of 316~mW during inference for the pruned and quantized model, with an inference duration of 27~ms.
In comparison, the IoT device has a baseline power consumption of 300~mW when not performing inference. Consequently, the additional power consumption during inference is relatively small.
This efficiency is critical for extending the device’s operational lifetime, as the impact of each inference on overall energy consumption is minimal. By optimizing both the inference process and the model, energy usage can be further minimized, ensuring that the IoT device remains efficient and sustainable over time.

\section{Conclusion and future work}
Package monitoring is a critical challenge in the industry, both for operational efficiency and environmental sustainability. It appears that wake time and adaptation to different events throughout the package's lifecycle are key to extending the lifespan of IoT devices. To address this, we designed a system using convolutional neural networks capable of classifying events from accelerometer data while minimizing energy consumption and model size.

The main contributions include: the use of data augmentation algorithms tailored to imbalanced time series, an optimized CNN architecture for multi-class classification integrating adaptive thresholding to reduce errors, the application of compression techniques such as pruning and quantization to improve compatibility with resource-constrained microcontrollers, and a comprehensive implementation of the model on an ESP32 microcontroller, with energy consumption measurements, demonstrating efficiency in real-life scenarios.

The results are particularly promising: the compressed model achieves a size reduced by four (14~kB) with no significant loss in performance, maintaining a precision of 94.54\% for the "Forklift" class and 95.83\% for the "Truck" class. Furthermore, energy consumption measurements have a low power consumption, confirming that the proposed optimizations enable efficient deployment on IoT devices.


Future work includes exploring multimodal sensor fusion to enhance model performance.
Including new sensors will diversify input types and might give better performance results. 
With the same purpose, we plan to train the model with new unsupervised data, as for logistic reasons these data are easier to collect than supervised ones.

\bibliographystyle{IEEEtran}

\bibliography{cites}

\end{document}